\documentclass[10pt,twocolumn,letterpaper]{article}

\usepackage{cvpr}
\usepackage{times}
\usepackage{epsfig}
\usepackage{graphicx}
\graphicspath{{figs/}}
\usepackage{amsmath}
\usepackage{amssymb}
\usepackage{subcaption}

\def\mynote#1#2{{\it[#1:#2]\marginpar{\tiny{#1}}}}
\def\mynote#1#2{}

\makeatletter
\DeclareRobustCommand\onedot{\futurelet\@let@token\@onedot}
\def\@onedot{\ifx\@let@token.\else.\null\fi\xspace}
\def\eg{\emph{e.g}\onedot} 
\def\ie{\emph{i.e}\onedot}

\def\etal{\emph{et al}\onedot}
\makeatother

\DeclareMathAlphabet\mathbfcal{OMS}{cmsy}{b}{n}

\cvprfinalcopy 

\def\cvprPaperID{527} 

\ifcvprfinal\fi
\begin{document}

\title{Becoming the Expert - Interactive Multi-Class Machine Teaching}

\author{Edward Johns \qquad Oisin Mac Aodha \qquad Gabriel J. Brostow \\
University College London\\
{http://visual.cs.ucl.ac.uk/pubs/interactiveMachineTeaching}
}


\maketitle

\begin{abstract}
Compared to machines, humans are extremely good at classifying images into categories, especially when they possess prior knowledge of the categories at hand.
If this prior information is not available, supervision in the form of teaching images is required.
To learn categories more quickly, people should see important and representative images first, followed by less important images later -- or not at all. 
However, image-importance is individual-specific, i.e. a teaching image is important to a student if it changes their overall ability to discriminate between classes. 
Further, students keep learning, so while image-importance depends on their current knowledge, it also varies with time.

In this work we propose an Interactive Machine Teaching algorithm that enables a computer to teach challenging visual concepts to a human. 
Our adaptive algorithm chooses, online, which labeled images from a teaching set should be shown to the student as they learn. 
We show that a teaching strategy that probabilistically models the student's ability and progress, based on their correct and incorrect answers, produces better `experts'. 
We present results using real human participants across several varied and challenging real-world datasets.
\end{abstract}

\section{Introduction}


Large, manually annotated image datasets have contributed to recent performance increases in core computer vision problems such as object detection and classification~\cite{everingham2010pascal,ILSVRCarxiv14,MSRcoco}. 
In cases where the visual categories of interest are generic everyday objects, annotation can be completed by crowd sourcing labels from the internet using services such as Mechanical Turk~\cite{mTurk}. 
A typical image labeling task begins with a set of instructions to the annotator, showing them example images from the classes of interest. 
The annotator is then asked to assign class labels to new images where the ground truth is unknown. 

But what happens if the annotator is unsure? 
This is a real problem when annotators are incorrectly assumed to have prior knowledge of the classes of interest from which they can generalize, either from everyday life or from specialized training. 
For many problems, highly specialized, domain specific knowledge, acquired through extensive training, is needed before someone can differentiate between potentially multiple, highly self-similar object categories.

\begin{figure}[t!]
  \centering
  \includegraphics[width=\linewidth]{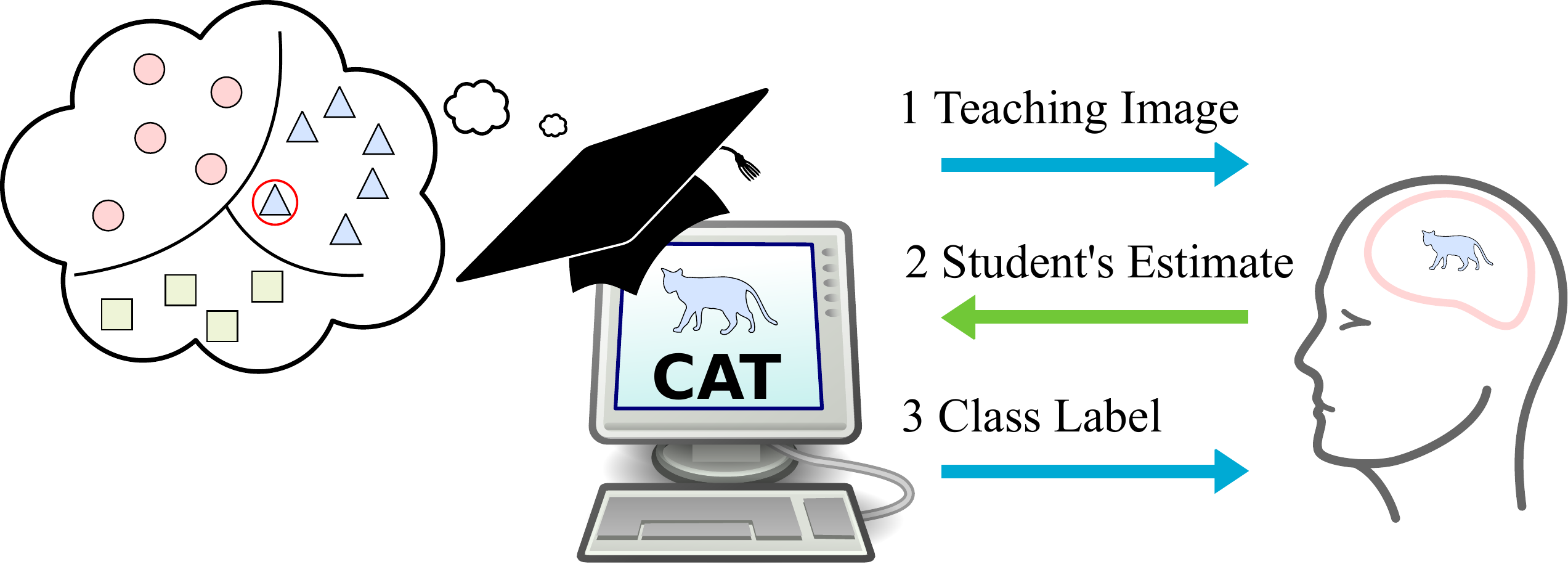}
  \caption{In Interactive Machine Teaching the computer teaches the human learner, one image at a time. It begins by showing them an image from the larger labeled image dataset while concealing the true class label. The learner/student responds with their estimate of the image's class. The teacher then updates their model of the student, and finally reveals the correct answer to them. This process is repeated with further images until teaching ends.}
  \label{teaching_overview}
\end{figure}

Designing the set of teaching images or `teaching set' to show the annotators is challenging, because each annotator will have a different degree of expertise. 
While it is possible to model the uncertainty and noise generated from groups of annotators to improve their collective performance~\cite{raykar2009supervised,welinder2010multidimensional,long2013active}, these approaches tend to downweight votes from weak annotators by learning to trust the experts. 
In this paper, we pose the  question - how does one become an expert? 
We posit that a human's discriminative ability for a given visual classification task can be improved by better modeling the teaching process required to make them experts.

The family of methods referred to as Machine Teaching offers a general solution to the problem of teaching humans~\cite{zhu2015machine,goldman1995complexity, zhu2013machine,singla2014near,teachingNIPS2014}.
Machine Teaching is not the same as Active Learning~\cite{Settles12activelearning}. 
In Active Learning, the computer's goal is to learn more accurate models given the smallest amount of supervision.
This is achieved by carefully selecting only the most informative datapoints to be labeled by the human. 
In Machine Teaching, the computer, rather than the human, is the (perfect) oracle and is tasked with delivering a teaching set to the human student to help them learn the given task more effectively.
Teaching the human a new skill is useful in its own right. Further, they are now better positioned to accurately annotate the additional unlableled data outside the original teaching set. Automatic teaching algorithms have applications in many domains from education, language learning, medical image analysis, biological species identification~\cite{aodha2014putting}, and more. 
Crucially, for automated teaching to be effective, it needs to be able to assess the student's current knowledge, and have a mechanism for selecting teaching examples to best improve this knowledge.


In this work, we focus on the task of image classification. 
Here, it is not possible for the teacher to directly `teach' the high-dimensional decision boundary to the human learner, so instead, the student must learn this boundary by being shown teaching images. 
Our goal during teaching is to choose teaching images that will maximize the student's classification ability in the minimum amount of teaching time. 
Unlike computers, humans have both limited and imperfect memory for instance-level recognition, especially during the initial learning of a task~\cite{giguere2013limits}.
However, humans have the advantage of possessing the ability to generalize to unknown examples and perform domain adaptation given only few instances.
The majority of previous work in Machine Teaching has focused on non-interactive teaching, where one teaching set is computed offline, independent of feedback from each student~\cite{singla2014near,teachingNIPS2014}. 
In this work, we address the under-explored problem of interactive teaching~\cite{du2011active, basu2013teaching}.
Here, the teacher can adapt their teaching set online, based on the current performance of the individual student (see Figure~\ref{teaching_overview}).




We propose an algorithm that interactively teaches multiple visual categories to human learners. 
Our contributions are threefold: 
1) Unlike computers, humans are not optimal learners.
Our algorithm models student ability online, resulting in teaching sets that are adapted to each individual student. 
We make no assumptions regarding the internal learning model used by the students.
Instead, we present them with teaching images that attempt to reduce their predicted future uncertainty based on an estimate of their current knowledge.
2) Our teaching algorithm reduces the amount of time it takes students to learn categorization tasks involving multiple classes. 
Experimentally we show that real human participants, using our algorithm, perform better than other baselines on several challenging datasets.
3) Finally, we provide a web based interface and framework for exploring new teaching strategies. 
Our intention is that this will encourage the development of new and diverse teaching strategies for a variety of human visual learning tasks.

\section{Related Work}
Here we cover the most closely related work in Machine Teaching. 
As we are concerned with the task of image categorization, we focus on research concerning teaching classification functions. However, it is worth noting that different types of teaching tasks have been explored in the literature, \eg sequential decision tasks~\cite{cakmak2012algorithmic}. 
How humans acquire and represent categories is an active area of research in visual psychology. 
Many candidate models for category acquisition and representation in humans exist, and for an overview we direct the readers to~\cite{love2013, richler2014visual}.
In this work, our goal is not to model these internal processes directly, but to instead treat the human as a stochastic black box learner.
For convenience, we divide the related work in Machine Teaching into two areas - batch (fixed) teaching and interactive (adaptive or online) teaching. 
For a recent, and general, introduction to Machine Teaching, please see~\cite{zhu2015machine}.

\subsection*{Machine Teaching - Batch (Fixed)} 
In batch-based teaching, the teacher's goal is to construct an optimal set of teaching examples offline, which are then presented to the student during teaching.
Early work in this area focused on the theoretical analysis of the teaching dimension~\cite{goldman1995complexity}.
The teaching dimension is defined as the minimum number of examples required from a given concept to teach the concept to a student.
Like many other works in teaching,~\cite{goldman1995complexity} makes the simplifying assumption that the student has perfect memory (\ie once shown an example the student will remember it in the future) - an assumption that is violated in real world teaching. 
Other theoretically motivated works, while interesting, provide little validation on real human subjects~\cite{balbach2009recent, doliwa2010recursive, zilles2011models}.

More recently, Zhu~\cite{zhu2013machine} attempted to minimize the joint effort of the teacher and the loss of the student by optimizing directly over the teaching set. 
The proposed noise-tolerant model assumes that the student's learning model is known to the teacher, and that it is in the exponential family. 
In follow-on work, Patil~\etal\cite{teachingNIPS2014} maintain that unlike computers, which have infinite memory capabilities, humans are limited in their retrieval capacity. 
Motivated by real human studies~\cite{giguere2013limits}, they show that modeling this limited capacity improves human learning performance on tasks involving simple one-dimensional stimuli. 

Most related to our work, Singla~\etal\cite{singla2014near} teach binary visual concepts by showing images to real human learners. 
Their method operates offline and tries to find the set of teaching examples that best conveys a known linear classification boundary. Experiments with Mechanical Turkers show an improvement compared to other baselines, including random sampling. 
Their approach attempts to encode some noise tolerance into the teaching set, but is still unable to adapt to a student's responses online during teaching, because the ordering of the teaching images is fixed offline.

\subsection*{Machine Teaching - Interactive (Adaptive)} 
Real human students are often noisy, especially in the early stages of learning when the concepts to be learned are not formed in their minds. 
Additionally, students do not all learn at the same rate - concepts that are difficult for some students may be easier for others.
In Interactive Teaching (Figure~\ref{teaching_overview}), the teacher receives feedback from the student as teaching progresses. 
Given this feedback, the teaching strategy can adapt to the current ability of an individual student over time.

Using a probabilistic model of the student and a noise-free learning assumption, Du and Ling~\cite{du2011active} propose a teaching strategy called `worst predicted'.
This strategy is similar to uncertainty sampling, which is commonly found in Active Learning~\cite{Settles12activelearning}. 
However, unlike Active Learning, in Machine Teaching the teacher has access to the ground truth class labels and can use this to assess the student's performance during teaching. 
Experimentally, we show that their strategy performs sub-optimally as it only seeks to show the student the image that they are currently most uncertain about, without regard for how informative that image may be in relation to others. 
As a result, it is very susceptible to teaching outliers, \ie unrepresentative images at the fringes of the teaching set. 

In one of the few interactive teaching papers that deal with visual concepts, Basu and Christensen~\cite{basu2013teaching} evaluate human learning performance in binary classification using three different teaching methods. 
Students were tasked with classifying simple synthetically generated (and linearly separable) depictions of mushrooms into one of two categories. 
They do not explicitly model labeling noise from the student, but instead investigate different interface designs and feature space exploration methods to help teach the students. 

In this paper, we address the problem of interactive multi-class teaching with real images by directly modeling the student's ability as they provide feedback during teaching.





\newcommand{\Prob}   {\ensuremath{P\xspace}}
\newcommand{\TeachingSet} {\ensuremath{\mathcal{D}_t}}
\newcommand{\LabelledData} {\ensuremath{\mathcal{D}_l}}
\newcommand{\UnlabelledData} {\ensuremath{\mathcal{D}_u}}
\newcommand{\EstLearnerProb} {\ensuremath{\hat{P}_l\xspace}}
\newcommand{\LearnerProb} {\ensuremath{P_l\xspace}}
\newcommand{\EstOutProb} {\ensuremath{\hat{P}_{\LabelledData}\xspace}}
\newcommand{\Error}   {\ensuremath{\mathcal{E}}}
\newcommand{\Loss}   {\ensuremath{L\xspace}}
\newcommand{\vx}   {\ensuremath{\mathbf{x}}}
\newcommand{\Indicator}  {\ensuremath{\mathrm{I}}}
\newcommand{\argmin}{\arg\!\min}
\newcommand{\argmax}{\arg\!\max}
\section{Machine Teaching}
In this section we formally define our Machine Teaching task. 
Our teacher-computer has access to a labeled dataset $\mathcal{D} = \left\{(\mathbf{x}_1, y_1), ..., (\mathbf{x}_N, y_N)\right\}$ where each $\vx_i$ is an $M$ dimensional feature vector encoding an image $I_i$, and $y_i\in \left\{1, ..., C\right\}$ is its corresponding class label.
The teacher's goal is to `teach' the classification task to the human learner by showing them images from the dataset $\mathcal{D}$.
We refer to these teaching images as the `teaching set', ${\TeachingSet}$, a subset of images from $\mathcal{D}$ where $|{\TeachingSet}| \ll |\mathcal{D}|$. 
In each round of interactive teaching, the teacher first selects an image represented by the feature vector $\vx_t$ to show to the human learner.
The teacher displays images to the students as it is not possible to directly show them the high dimensional feature vector $\vx_t$.
The selection of the image to show is based on a process we refer to as the `teaching strategy', $\mathcal{S}$, used by the teacher. 
First, the teacher only shows the image and does not yet display the ground truth class label. 
By not revealing the class label, the teacher is able to ask the student to state which class they believe the image belongs to. 
After receiving the student's response, the teacher then updates its model of the student, and then reveals the ground truth label.
Teaching proceeds for a set number of teaching rounds, and during each iteration, the teacher acquires a better understanding of the student's current ability. 
Figure~\ref{teaching_overview} outlines one teaching iteration.

With access to ground truth, the teacher trivially knows the conditional distribution $\Prob(y_i\,|\,\vx_i)$ for each datapoint $\vx_i$.
The student learner has a corresponding distribution $\LearnerProb(y\,|\,\vx)$, based only on training examples they have seen so far.
During teaching, the teacher seeks to minimize the student's expected loss 
\begin{equation}\label{learnLoss}
\mathbb{E}_\mathbf{x} = \Loss\left( \Prob(y\,|\,\vx), \LearnerProb(y\,|\,\vx) \right),
\end{equation}
over the dataset, where $\Loss()$ is an appropriate classification loss function. 
However, the teacher has no way of directly observing the student's true class conditional distribution, $\LearnerProb(y\,|\,\vx)$, so instead must approximate it as $\EstLearnerProb(y\,|\,\vx)$.
In this paper, we represent $\EstLearnerProb(y\,|\,\vx)$ using a probabilistic, semi-supervised, classifier.

\subsection{Teaching Strategies}\label{teachingStrt}
The optimal teaching strategy is the one that minimizes the student's expected loss from Equation~(\ref{learnLoss}).
A simple strategy for choosing the next teaching image is to randomly sample from the dataset $\mathcal{D}$. 
Random sampling ($\mathcal{S}_{rnd}$) does not model the student and is therefore unable to adapt to their ability. 
This lack of adaptation can manifest itself in two ways - 1) redundantly presenting teaching examples of concepts that have already been learned by the student, and 2) not directly reinforcing concepts that the student has shown themselves (through feedback) to be uncertain about.

Du and Ling~\cite{du2011active} proposed a strategy called `worst predicted', here $\mathcal{S}_{wp}$, which is  related to uncertainty sampling commonly used in Active Learning~\cite{Settles12activelearning}. 
However, unlike in Active Learning, in Machine Teaching, the computer does have access to the ground truth labels.
Their strategy selects the next teaching image as the one whose prediction deviates most from the ground truth,
\begin{equation}\label{eqEE2}
\vx_t = \argmin_{\vx} \EstLearnerProb(\bar{y}\,|\,\vx),
\end{equation}
where $\bar{y} = \argmax_{y} \Prob(y|\vx)$ is the ground truth class label known to the teacher. 
The disadvantage of this approach is that it is prone to proposing outliers as teaching images, as they tend to be highly uncertain under the current model. 
One potential solution to this problem is to weight the datapoints by some measure of local density in the feature space~\eg\cite{Settles12activelearning, ebert2012ralf}. 

\subsubsection{Expected Error Reduction Teaching}
Our teaching strategy, which we refer to as $\mathcal{S}_{eer}$, takes inspiration from optimal sampling methods found in Active Learning~\cite{roy2001, Zhu03combiningactive, MacAodhaCVPR2014}.
Unlike $\mathcal{S}_{wp}$, $\mathcal{S}_{eer}$ chooses the teaching image which, if labeled correctly, would have the greatest reduction on the future error over the images that are not in the teaching set, $\UnlabelledData = \mathcal{D} \setminus \TeachingSet$, where
\begin{equation}\label{eqEER}
\vx_t = \argmin_{\vx_p} \sum_{\vx_i,\bar{y}_i \in  \UnlabelledData}(1 - \EstLearnerProb^{+(\vx_p, \bar{y}_p)}(\bar{y}_i\,|\,\vx_i)).
\end{equation}
Here, $\EstLearnerProb^{+(\vx_p, \bar{y}_p)}$ is the updated estimate of the student's conditional distribution if they were shown $\vx_p$ and in turn labeled it correctly.
This strategy has the advantageous property that it first concentrates on regions of high density in the feature space, and as the student improves, refines the boundaries between these regions. 
In the context of Active Learning, this is referred to as the exploration versus exploitation trade off.
This is related to the approach to learning advocated by curriculum learning, which focuses on easy concepts first and progressively increases the difficulty~\cite{bengio2009curriculum}.


\subsection{Modeling the Student}
In this work we approximate the student's conditional distribution given the teaching set, $\EstLearnerProb(y\,|\vx, \TeachingSet)$, using graph based semi-supervised learning~\cite{zhou2004learning, zhu2003semi}. 
Using the Gaussian Random Field (GRF) semi-supervised method of~\cite{zhu2003semi}, we can propagate the student's estimate of the class labels for the current teaching set, $\TeachingSet$, to the unobserved images $\UnlabelledData$ by defining a similarity matrix $W \in \mathbb{R}^{N\times{}N}$.
The benefit of using a graph based approach is that we do not need to work directly in feature space, and can instead use the similarity, $w_{ij}$, between image pairs. 
This gives us the flexibility of allowing similarity to be defined using feature vectors extracted from the images, human provided attributes, or using distance metric learning~\cite{kulis2012metric}.

If we are given a feature representation for our teaching set, one common approach for computing the similarity $w_{ij}$ between two images uses an RBF kernel
\begin{equation}\label{graph_sim}
w_{ij} = \exp(-\gamma{}\lVert \mathbf{x}_i - \mathbf{x}_j \rVert{}_2^2),
\end{equation}
where $\gamma{}$ is a length scale parameter that controls how much neighboring images influence each other.  
Using matrix notation of~\cite{zhu2003semi}, we define an $N\times C$ matrix $F = \EstLearnerProb(y\,|\,\vx, \TeachingSet)$, where each element $f_{ic} = \EstLearnerProb(y_i = c\,|\,\vx_i)$. 
We can propagate information from the labels provided by the student for the teaching set, encoded as a $|\TeachingSet|\times C$ matrix $F_{t}$, to the unlabeled images $F_u$,
\begin{equation}
F_u = (S_{uu} - W_{uu})^{-1}W_{ut}F_{t},
\end{equation}
where $S$ is a diagonal matrix with entries $s_{ii} = \sum_j w_{ij}$. 
All entries in $F_{t}$ are $0$, except where the human learner has estimated (correctly or incorrectly) the class label $c$ for teaching image $\mathbf{x}_{i}$, which we set to $f_{ic} = 1$. 
$W_{uu}$ is the similarity matrix for the unobserved images, a subset of the full matrix $W$. 
As in~\cite{zhu2003semi}, we can efficiently evaluate Equation (\ref{eqEER}) using standard matrix operations for datasets featuring $2000$ images in under one second using unoptimized Python code.



\section{Experiments}
To validate our proposed multi-class teaching strategy, we performed studies on real human subjects.
Participants were recruited through Mechanical Turk~\cite{mTurk}, and interacted with our system remotely using our custom made web interface, built using the Python-based Django web framework~\cite{django}.

\begin{figure*}[t!]
  \centering
  \includegraphics[width=\linewidth]{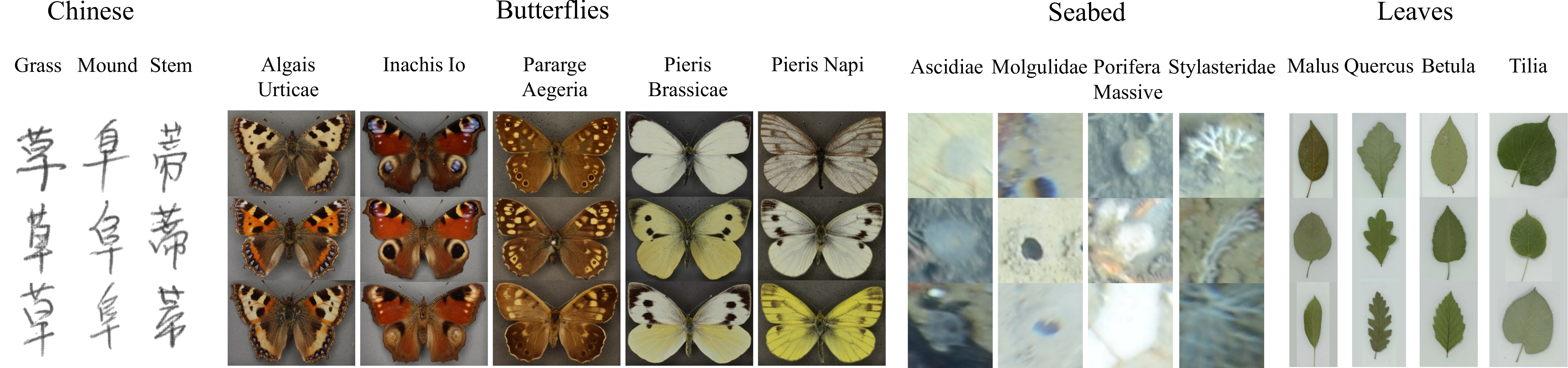}
  \caption{Example images from the four datasets used in our experiments. Each column shows three random images per class. Note, that these images are challenging to categorize as they exhibit a large amount of intra-class variation. Additionally, the `Seabed' images are particularly difficult as they were captured `in the wild' and contain occlusion and clutter.}
  \label{datasetFigure}
\end{figure*}

\subsection{Data}
For our experiments, we selected four different datasets, summarized in Table~\ref{datasetTable}.
To ensure that the teaching tasks were challenging to participants and one-shot learning was not possible, we chose datasets with small inter-class variation and large intra-class variation. 
Example images from each of the classes are presented in Figure~\ref{datasetFigure}.
Unlike standard classification datasets featuring everyday objects \eg~\cite{griffin2007caltech, everingham2010pascal}, our datasets contain image categories that are challenging for non-domain experts to discriminate between, as they are made up of uncommon classes. 

\begin{table}[b!] 
 \centering 
 \begin{tabular}{l|ccc} 
 \hline 
\bf{Dataset}  & \bf{\# Classes} & \bf{\# Images per Class} & \bf{Origin} \\ \hline
Chinese & 3 & 237-240 & \cite{liu2011casia}\\ 
Butterflies & 5 & 300 & -\\ 
Seabed & 4 & 100 & -\\ 
Leaves & 4 & 102-150 & \cite{kumar2012leafsnap}\\ 
\hline 
 \end{tabular}
 \caption{Summary of the datasets used, showing the number of classes and the minimum and maximum number of images per class.}
 \label{datasetTable}
\end{table}

Two of the datasets, `Butterflies' and `Seabed', were collated by the authors of this paper from ongoing scientific studies into visual species identification. 
`Butterflies' is a subset of a larger collection of British butterfly images from a museum collection captured over a period of 100 years.
`Seabed' is a set of images of underwater species taken from a study attempting to measure the effects of trawling on underwater bio-diversity. Both datasets were curated and annotated by domain experts.  

Image features were extracted using the publicly-available ConvNet system of~\cite{jia2014caffe}. 
For each dataset, we computed features using a network pre-trained on the ImageNet 2012 challenge dataset~\cite{ILSVRCarxiv14}.
We then fine-tuned the fully connected layers using the known ground truth class labels for each of our datasets, which produced a separate ConvNet for each dataset.
To construct the similarity matrix $W$ of (\ref{graph_sim}), we reduced the dimensionality of the ConvNet features from $4096$ to $50$ using PCA, and set the length scale parameter $\gamma$ to $0.025$ for all datasets.
In our initial experiments, we explored custom-designed HoG-based features~\cite{dalal2005cvpr} which we found to perform worse compared to our fine-tuned ConvNet. 
Here, the additional supervised information provided during fine-tuning produces a representation where images from the same class are more smoothly distributed in feature space.
A feature space that is better aligned with the student's view of similarity should benefit all probabilistic strategies equally. 
It would also be possible to compute similarity between teaching images by crowd-sourcing image rankings from a set of users, \eg~\cite{o2014exploratory}.
However, we found our ConvNet features to be a good balance between reducing the amount of additional supervised information required for each teaching task and real students' performance.
Code and data are available on our project website.



\subsection{Experimental Design}
To evaluate our teaching algorithm, we conducted experiments on participants recruited through Mechanical Turk~\cite{mTurk}.
Previously, Crump~\etal\cite{crump2013evaluating} have shown that it is possible to replicate results from classic category learning experiments using Mechanical Turk.
Using a similar experimental setup to~\cite{medin1978context}, our participants were first presented with a sequence of teaching images, which were then followed by a sequence of testing images.

For each experiment, participants were first told how many classes they were being asked to learn. 
Then teaching commenced using the interactive teaching loop illustrated in Figure~\ref{teaching_overview}.
For each teaching image, participants were first shown the image, asked to estimate its class label by clicking on the corresponding button in our web interface, and then provided with the correct answer. 
After receiving the estimated class label from the participant, the teaching strategy updates its model of the student and chooses the next image to be shown.
In contrast to the teaching phase, no corrective feedback in the form of the true class labels was provided in the testing phase.
The testing round was only used for evaluation purposes and is not necessary in real teaching scenarios. 
Test images were randomly chosen for each participant, with an equal from each class, and were excluded from the possible teaching set. 

Each participant was presented with a random dataset from Table~\ref{datasetTable}, combined with a random teaching strategy. 
For each dataset, the number of teaching images shown was set to three times the number of classes, and ten times the number for testing. 
In this way, the lengths of the teaching and testing rounds were proportional to the complexity of the task.
We experimented with longer teaching rounds ($> 40$ images) and testing at regular intervals between teaching images to achieve a learning curve.
However, we found through feedback that students became bored and frustrated with the enforced delay, encouraging them to drop out.

It is worth noting that our teaching tasks are significantly more difficult than most crowd-sourced image annotation tasks.
In typical annotation tasks, workers already possess strong prior knowledge of the concepts involved, whereas in our teaching tasks, the participants were unlikely to have prior domain expertise.
We surveyed participants at the start of the task to ensure that they possessed no prior task knowledge and we rejected results for those who claimed to have even moderate familiarity of any of the classes. As such, the student's answer to the first teaching image was always a random guess.
To avoid workers who were seemingly clicking at random, we also rejected results for those whose average response time per image was too fast ($<3$ seconds) during testing.
To encourage a conscientious effort in learning, we paid workers a bonus if they scored higher than a threshold during testing. 
After discarding noisy participants, we collected results from between $25$ and $35$ participants per strategy/dataset combination.

\subsection{Baseline Strategies}
In addition to the baseline teaching strategies outlined in Section~\ref{teachingStrt}, we also compared to two other baselines $\mathcal{S}_{cc}$ and $\mathcal{S}_{batch}$.
For $\mathcal{S}_{cc}$, or class centroids, we computed the feature space centroids for each class for a given dataset, and students were only presented with the images represented by these centroids during teaching. 
Teaching images were selected by randomly choosing from one of these centroids. 
If there was little intra-class variation, if one-shot learning was possible, or if the classes were familiar to the student, we would expect this baseline to perform very well.
The final baseline, $\mathcal{S}_{batch}$, is similar to offline batch teaching algorithms such as~\cite{singla2014near}. 
Here, the ordering of the teaching images was computed offline.
We computed the ordering using the $\mathcal{S}_{eer}$ algorithm, but assuming that if shown an image, the student would always label it correctly. 
Given this assumption, the selection of teaching images is deterministic and is identical for all students regardless of their responses.
Recent strategies for offline binary teaching, such as~\cite{singla2014near}, are not directly applicable for comparison because we operate in the challenging interactive multi-class classification scenario.

\subsection{Human Experiments}
\begin{figure*}[t!]
\centering
\setlength{\tabcolsep}{1pt}
\begin{tabular}{cccc}
\includegraphics[width=0.25\linewidth]{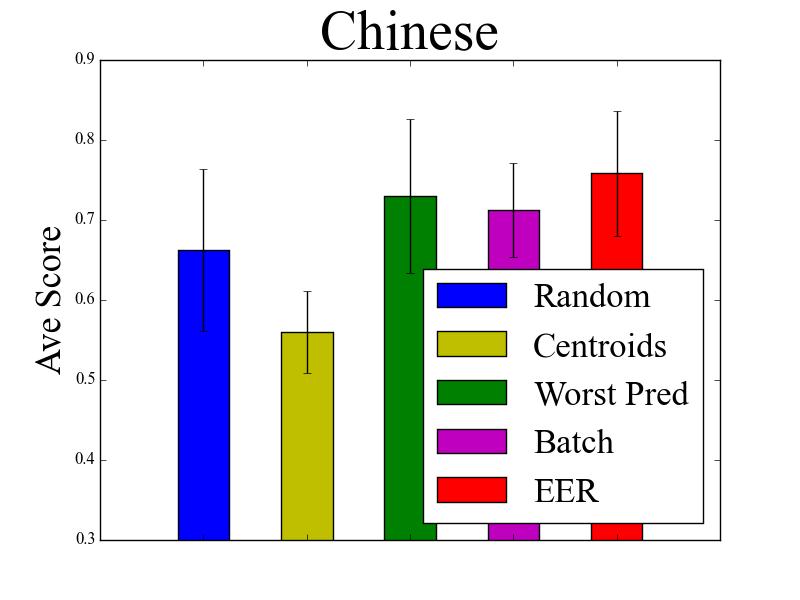} &
\includegraphics[width=0.25\linewidth]{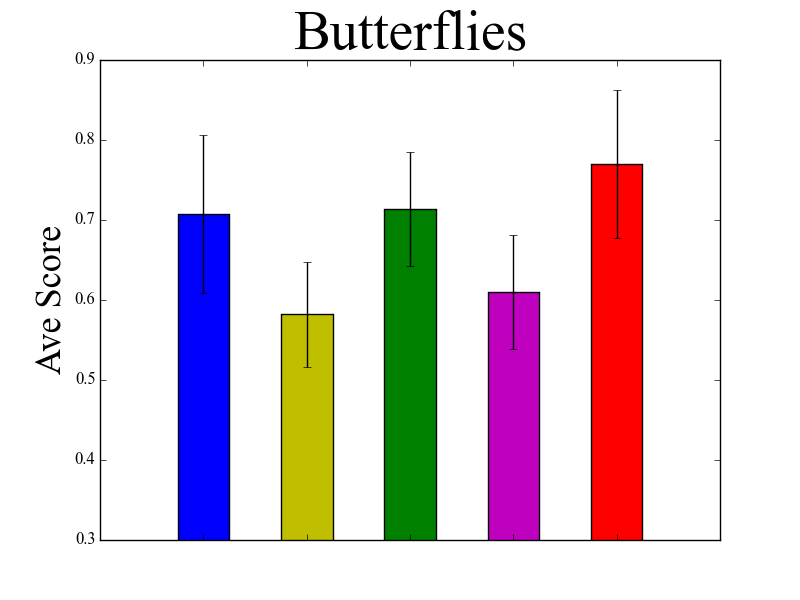} &
\includegraphics[width=0.25\linewidth]{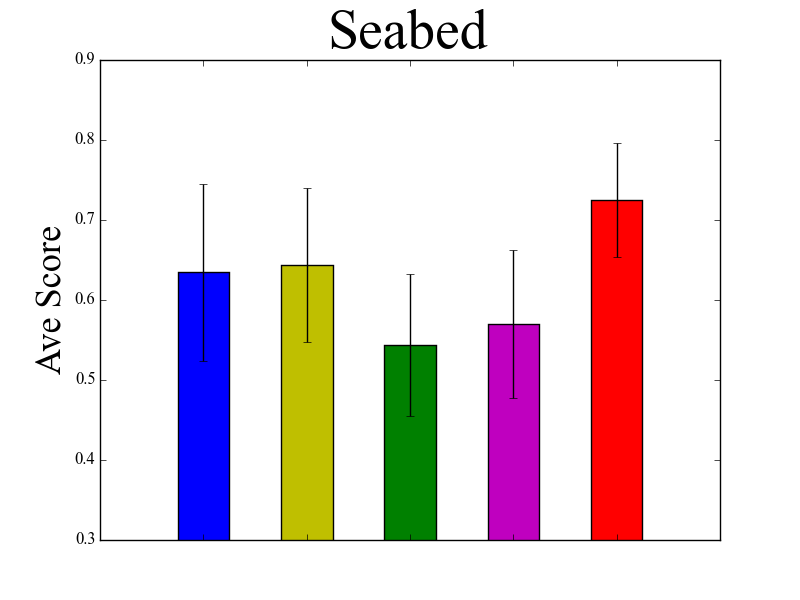} &
\includegraphics[width=0.25\linewidth]{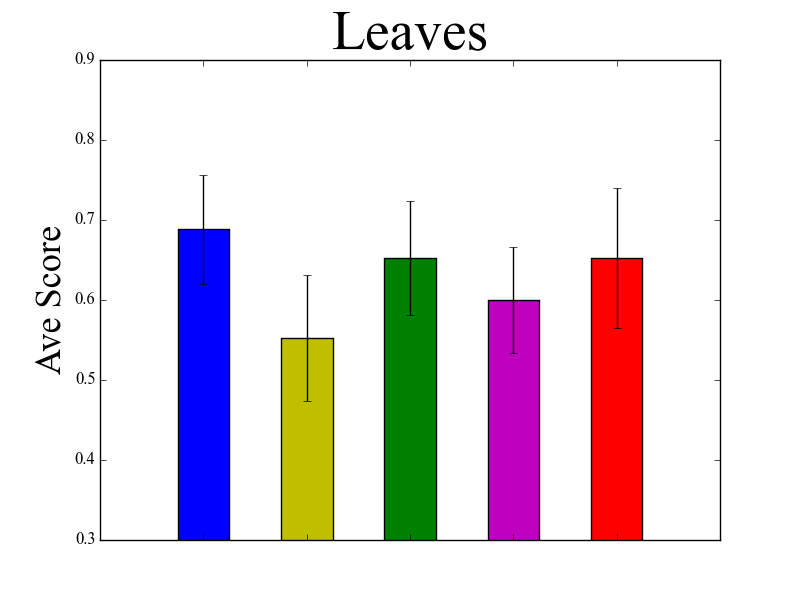} \\
\end{tabular}
\vspace{-12pt}
\caption{Human experiment results across the four datasets described in Table~\ref{datasetTable}, showing the average scores after the testing phase across all participants. Human participants on Mechanical Turk using our Expected Error Reduction based teaching strategy (here EER) tend to have better recognition performance on average after teaching, compared to the other baselines.}
\label{results_curves_real}
\end{figure*}

Results from human participants are summarized in Table~\ref{timingResults}. Results for individual datasets are depicted in Figure~\ref{results_curves_real}, where the average number of testing images answered correctly are shown for each dataset and strategy combination. 
We can see that our $\mathcal{S}_{eer}$ method outperforms the other teaching strategies on the `Chinese', `Butterflies', and `Seabed' datasets. 
In these three, our method is consistently the best performing, while the other methods vary in performance depending on the specific dataset. 
As we can see from Table~\ref{timingResults}, there is no clear `second-best' method, and the offline $\mathcal{S}_{batch}$ and uncertainty  $\mathcal{S}_{wp}$ strategies are often outperformed by random  $\mathcal{S}_{rnd}$. 
$\mathcal{S}_{eer}$'s performance is most pronounced on the `Seabed' dataset, which also contains the most haphazard images, due to the acquisition of data from cameras `in the wild', as opposed to neatly-framed imaging in controlled laboratory conditions.

Average timings during testing for the different strategies, calculated as the time between being shown a test image and submitting an answer, are presented in Table~\ref{timingResults}.
Participants taught using our method tend to answer more quickly compared to the other strategies.
$\mathcal{S}_{rnd}$ and  $\mathcal{S}_{cc}$ also have low response times, but students' poorer performance at test time possibly indicates a level of false-confidence. 

\begin{table}[t!] 
 \centering 
 \begin{tabular}{ll|c|c} 
 \hline
 \multicolumn{2}{c|}{\bf{Strategy}}    & \bf{Ave. Time (ms)} & \bf{Ave. Score}\\ \hline
Random & $\mathcal{S}_{rnd}$ & 4876 & 0.67\\ 
Centroids & $\mathcal{S}_{cc}$  & 4706 & 0.58 \\ 
Worst Pred. & $\mathcal{S}_{wp}$& 5237 & 0.66\\ 
Batch & $\mathcal{S}_{batch}$& 6216 & 0.64\\ 
\bf{EER (Ours)} & $\mathbfcal{S}_{eer}$ & \bf{4659} & \bf{0.73}\\
\hline 
 \end{tabular}
 \caption{Average participant response times during testing, and test set scores across all datasets.}
 \label{timingResults}
\end{table}

Table ~\ref{pTest} provides p-values for the statistical significance of our results. 
Two-tailed tests were conducted with a null hypothesis that the distributions of scores for our method across all datasets, and the competing method, are statistically similar, based on a Gaussian assumption. 
The p-values obtained are well within the standard measure of $0.05$ for testing statistical significance, indicating that our results are not due to chance.

\begin{table}[b!] 
 \centering 
 \begin{tabular}{ll|r} 
 \hline
 \multicolumn{2}{c}{\bf{Strategy}}    & \bf{P-value} \\ \hline
Random & $\mathcal{S}_{rnd}$ & 0.0138\\ 
Centroids & $\mathcal{S}_{cc}$  & $<$ 0.0001\\ 
Worst Pred. & $\mathcal{S}_{wp}$& 0.0027\\ 
Batch & $\mathcal{S}_{batch}$&  $<$ 0.0001\\ 
\hline 
 \end{tabular}
 \caption{Two-tailed p-values for hypothesis tests on the statistical significance of our method compared to all others.}
 \label{pTest}
\end{table}

\begin{figure}[h]
  \centering
  \includegraphics[width=\linewidth]{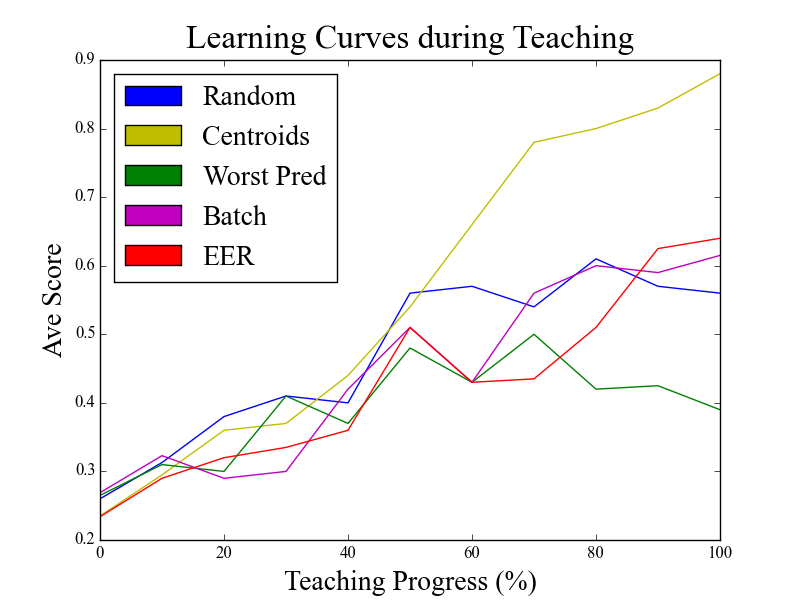}
  \caption{Average learning curves across all students and datasets during the teaching phase.}
  \label{teaching_curve}
\end{figure}

Figure~\ref{teaching_curve} shows the average learning curves for the five teaching strategies obtained during teaching. 
The average score for each $10\%$ progress interval (through the training set) is calculated by averaging the number of correct responses over all students and datasets at that point along the teaching phase. 
Note that this is not equivalent to the true learning curve, as images are chosen to actively teach the student,
rather than to assess a snapshot of their performance.
We see a general trend of improving recognition rates with further teaching images. 
However, $\mathcal{S}_{cc}$ gives a false sense of performance because the same centroid images are repeatedly shown, thus the student overfits to these images and typically fails to generalize during testing. 
Unlike the others, the
uncertainty based $\mathcal{S}_{wp}$ strategy has a relatively flat learning curve, because the outlier images shown are challenging to learn. 
This underfitting gives students only a weak understanding of each class's variability.

\begin{figure*}[t!]
  \centering
  \includegraphics[width=0.9\linewidth]{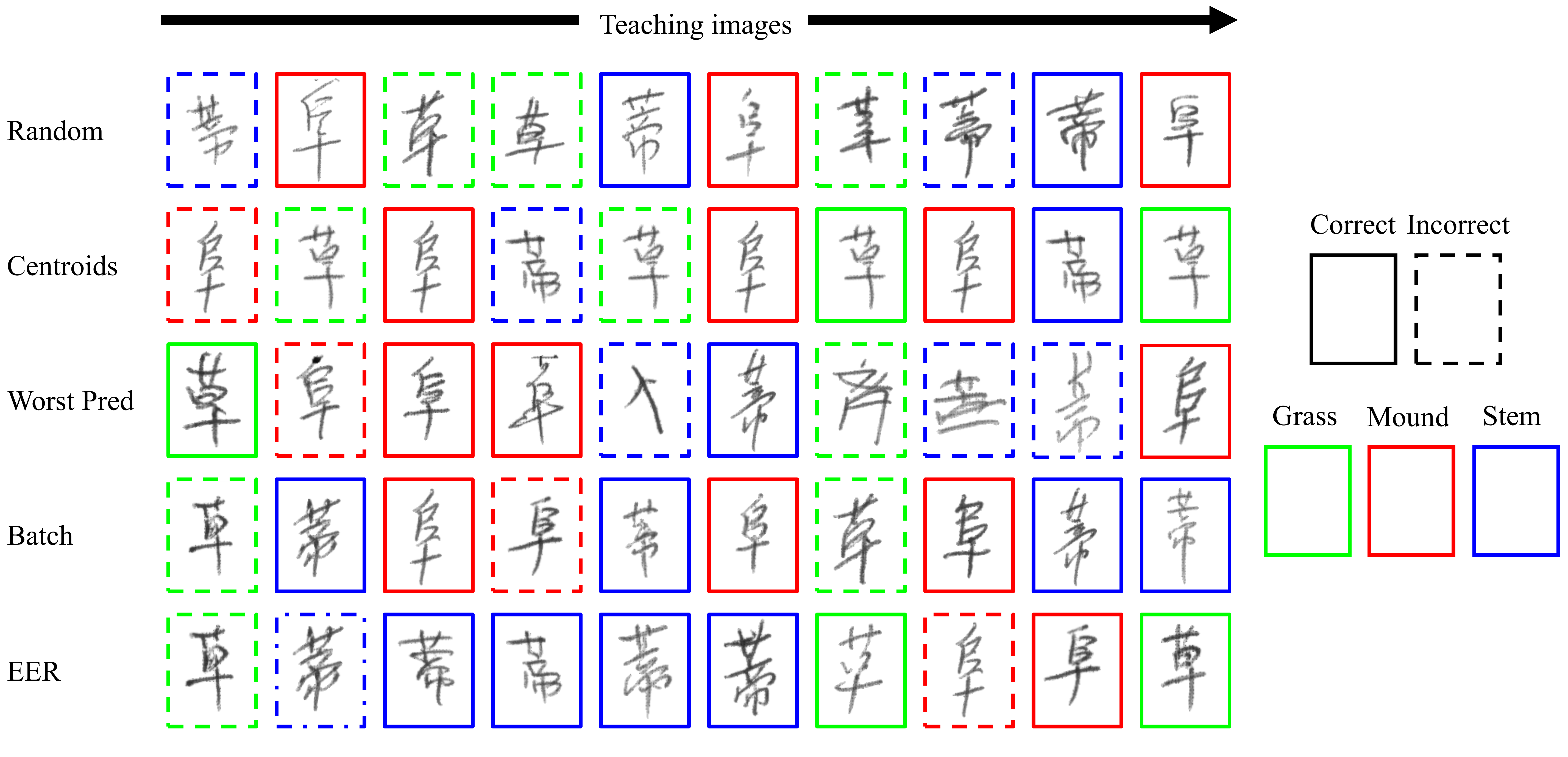}
  \vspace{-5pt}
  \caption{Example images and responses for the different teaching strategies from $5$ sample individual students during teaching of the 'Chinese' dataset. Solid boxes indicate correct answers, dashed lines for incorrect answers, and box colors indicate the ground truth class labels.}
  \label{teaching_exs}
\end{figure*}

Figure~\ref{teaching_exs} shows examples of the teaching images shown to students for each of the five strategies with the 'Chinese' dataset. 
We see the capacity of $\mathcal{S}_{eer}$ to adapt to incorrect responses, where attention is given to the 'Stem' class due to an incorrect previous answer, before returning to teach 'Grass' due to its previous incorrect answer, and finally exploring the student's understanding of 'Mound'. 
On the other hand, $\mathcal{S}_{batch}$ is unable to adapt its teaching set and focuses on teaching 'Mound' and 'Stem' despite the student's poor performance with 'Grass'. 
$\mathcal{S}_{wp}$ begins by displaying reasonable examples, but ends by attempting to teach very unusual examples which are not representative of the dataset's distribution.

The performance of $\mathcal{S}_{eer}$ on the `Leaves' dataset shows an example where we do not perform better than the random baseline, but come joint second. 
A property unique to this dataset is the multi-modal nature of the leaves present, where each class in fact represents an entire genus, composed of a number of different species that do not all look the same. 
We found that human learners typically assumed unimodal distributions during teaching and would often focus on only a single species within the entire genus.

 

%
%

\subsection{Limitations}
Currently, our model does not attempt to directly recover from incorrect responses made by students in the past. 
If a student has previously given an incorrect answer, future teaching images will be selected from similar regions in the feature space.
However, earlier incorrectly labeled images will still influence the label propagation.
Allowing incorrectly labeled images to be relabeled could result in the teaching strategy continually presenting the same images until they are correctly labeled.
This behavior would be appropriate for a machine learner, but a human would quickly learn to cheat the learning task.
Any revision style strategy would have to be carefully designed to ensure that concepts that are already learned are not continually revisited due to a de-emphasizing of earlier teaching answers.

\section{Conclusion}
Machine Teaching has the potential to enable humans to learn concepts without human-to-human expert tutoring. 
By automatically adapting the curriculum to a student's ability and performance, teaching can be performed in situations where it is difficult or prohibitively costly to get direct access to domain-level expertise from a human teacher. 
In this work, we have taken a step in this direction by proposing an interactive multi-class teaching strategy. Its objective is to present to the student the teaching images that will be most informative, given an online estimate of their current knowledge.
Unlike other proposed strategies, we are less likely to teach outliers, and as a result, do not waste time showing unrepresentative images. 
Similar to curriculum learning~\cite{bengio2009curriculum}, our strategy initially focuses on representative images and then introduces more difficult ones over time, as the student's performance improves.

\subsection{Future Work}
Currently, we present teaching images to the students one at a time. 
In future, we plan to investigate different methods for displaying images. 
Visualizations such as pairwise comparisons~\cite{joshi2010breaking}, and highlighting local regions~\cite{deng2013fine} or parts~\cite{berg2013you}, may prove more effective at conveying discriminative details and characteristics of different categories. 
Some images are intrinsically more `memorable' than others~\cite{deza2015understanding, isola2013pami}, and incorporating such measures into teaching image selection may also improve test time performance.

In curriculum learning, task difficulty is increased as performance improves.
In future work, we shall also investigate other teaching paradigms such as the spiral approach to teaching~\cite{bruner1960process}.
In spiral learning, new categories are introduced over time while continually re-emphasizing the earlier concepts to ensure that they become committed to memory. 

Given that we can now teach humans visual categorization tasks in an automated fashion, in future work we intend to investigate what additional information we can extract from our students during and after teaching.
In contrast to machines, studies suggest that humans can learn with idealized versions of data that can have a different distribution from the test set~\cite{giguere2013limits}.
Exploring teaching as a domain adaptation problem could allow us to acquire annotations for data which is very different from our teaching set. 
Finally, we have assumed that our feature space is correlated with a student's concept of similarity.
It may be more effective to jointly estimate both the student's current ability and their notion of similarity during teaching.


\vspace{10pt}
\subsubsection*{Acknowledgements}
Funding for this research was provided by EPSRC grant EP/K015664/1 and by Sustainable Fisheries Greenland. We would like to thank Maciej Gryka for his web development advice, and the Natural History Museum London for the butterfly data. Edward Johns' funding was provided by the Greenland Benthic Assessment project at the Institute of Zoology at ZSL, in association with the Greenland Institute of Natural Resources.

{\small
\bibliographystyle{ieee}
\bibliography{active_teaching}
}

\end{document}